\documentclass{article}

 \usepackage[preprint]{neurips_2025}

\usepackage[utf8]{inputenc} 
\usepackage[T1]{fontenc}    
\usepackage{hyperref}       
\usepackage{url}            
\usepackage{booktabs}       
\usepackage{amsfonts}       
\usepackage{nicefrac}       
\usepackage{microtype}      
\usepackage{inconsolata}         

\usepackage{times}
\usepackage[inline]{enumitem}
\usepackage{latexsym}
\usepackage{amsmath}
\usepackage{amssymb}
\usepackage{multirow}
\usepackage{graphicx}
\usepackage{subcaption}
\usepackage{siunitx}
\usepackage{caption}
\usepackage[table]{xcolor}
\usepackage{geometry}
\usepackage{natbib}
\usepackage{wrapfig}

\newcommand{\todo}[1]{\textcolor{red}{\textbf{TODO:} #1}}
\newcommand{\ignore}[1]{}
\newcommand{\arsays}[1]{\textcolor{pink}{\textbf{AR: #1}}}

\title{Do Code Models Suffer from the Dunning-Kruger Effect?}

\author{
Mukul Singh$^{1}$\thanks{These authors contributed equally.} \quad Somya Chatterjee$^{1}$\footnotemark[1] \quad Arjun Radhakrishna$^{1}$ \quad Sumit Gulwani$^{1}$  \\
$^1$Microsoft\\
}

\begin{document}

\maketitle

\begin{abstract}
As artificial intelligence systems increasingly collaborate with humans in creative and technical domains, questions arise about the cognitive boundaries and biases that shape our shared agency. This paper investigates the Dunning-Kruger Effect (DKE), the tendency for those with limited competence to overestimate their abilities in state-of-the-art LLMs in coding tasks. By analyzing model confidence and performance across a diverse set of programming languages, we reveal that AI models mirror human patterns of overconfidence, especially in unfamiliar or low-resource domains. Our experiments demonstrate that less competent models and those operating in rare programming languages exhibit stronger DKE-like bias, suggesting that the strength of the bias is proportionate to the competence of the models. 
This aligns with human experiments for the bias.
We open source all benchmarks and predictions to encourage research in biases for AI models. These findings highlight the emergence of human-like cognitive biases in AI, raising new questions about trust and interpretability.
\end{abstract}

\section{Introduction}

Recent advances in artificial intelligence have led to models that not only automate complex tasks but also increasingly participate in creative and collaborative processes alongside humans, especially in coding tasks \cite{huynh2025large, cordeiro2024empirical, jelodar2025large, singh-etal-2023-codefusion, verbruggen2025executionguided}. As these systems become more integrated into domains such as art, design, and software engineering ~\cite{odeh2024comparative, anand2024comprehensive, chen2021evaluating, format5, singh2025diffusioncoderepairoperator, dutta-etal-2024-rar}, questions arise about the nature of human-machine symbiosis and the cognitive boundaries that separate or unite humans and AI.

A central aspect of this evolving relationship is the emergence of human-like cognitive biases within AI systems \cite{vakali2024rolling}, which has been studied extensively in numerous studies and policy discussions \cite{abrams2024addressing, vicente2023humans, landers2023auditing}. The Dunning-Kruger Effect (DKE), a well-documented phenomenon in psychology, describes how individuals with limited competence tend to overestimate their abilities \cite{mazor2021dunning, magnus2022statistical}. While DKE has been extensively studied in humans, its presence and implications in AI models remain underexplored, especially in contexts where machines are expected to collaborate, create, and self-assess.

In this work, we investigate whether large language models (LLMs) exhibit the Dunning-Kruger Effect in coding tasks. We argue that studying DKE in AI models is valuable for two reasons.
First, it offers a lens through which to examine model mis-calibration,
particularly in low-competence regimes.
Second, it allows us to test whether models exhibit human-like patterns of
overconfidence, which could have implications for trust, interpretability, and
downstream decision-making.

\ignore{
In this paper, we choose to study the Dunning-Kruger effect (DKE), a cognitive
bias observed in humans \cite{mazor2021dunning, magnus2022statistical} wherein
there is a mismatch between individuals' actual and self-perceived competence at
tasks.
More precisely, the lower an individual's performance on a task domain, the more
likely they are to overestimate their ability.
%
This phenomenon has profound implications for task performance and
decision-making in general, and specific mitigation strategies might be required
for individual domains.
We pick the domain of programming and the task of question-answering to study
this effect in AI models, and design controlled experiments to assess both
actual and self-perceived performance on these tasks.
Actual performance is measured using task completion accuracy, and perceived
performance through different absolute and relative confidence
measures~\cite{shrivastava2025language, elo1978rating, herbrich2006trueskill}.
}

Our results reveal that the models' perceived performance shows statistically
significant inflation compared to actual performance, similar to the effect
previously studied in humans.
The models' overestimation of their performance becomes more pronounced with
lower actual performance of the model and with increasing hardness of the tasks
(measured by rarity of the programming domain), aligning strongly with the
patterns observed in human cognition.
These findings underscore the importance of understanding cognitive biases in AI
systems and lay the groundwork for deeper interdisciplinary research at the
intersection of cognitive science and machine learning.

In this paper, we make the following contributions:(a) We provide statistically significant evidence of the Dunning-Kruger effect in AI models for coding tasks, (b) We analyze how the strength of this bias varies with (i) the model’s base performance and (ii) the rarity of the programming domain.

\section{Related Work}


\paragraph{Cognitive biases in AI models.} Studies have shown that LLMs can reflect human-like biases, including overconfidence and self-enhancement, despite lacking self-awareness  \cite{gu2024survey, salecha2024large, sun2025large, ye2024justice, aligned-code-gen}. These biases often stem from training data patterns or architectural choices \cite{geng2023survey, tjuatja2024llms}. Among these, overconfidence is particularly concerning, as it can lead to misleading outputs that appear authoritative but are incorrect, an issue that parallels the DKE \cite{dunning2003people, kruger1999unskilled, ehrlinger2008unskilled} observed in human cognition.

\paragraph{Generalization and confidence estimation}
In the context of code models, prior work has highlighted challenges in generalizing to rare programming languages
\cite{chen2024survey, cassano2024knowledge, giagnorio2025enhancing,
mora2024synthetic}.
While model accuracy drops on out-of-distribution tasks, confidence scores often remain high~\cite{chen2021evaluating}, revealing a disconnect between competence and self-assessment. Traditional confidence estimation methods, based on logits or self-reported probabilities, are frequently miscalibrated in unfamiliar domains
\cite{shorinwa2024survey, shen2024thermometer, yang2024can, li2024showing}.
Recent work introduces relative confidence estimation as a more robust alternative \cite{shrivastava2025language}.
These methods help uncover behavioral patterns like overconfidence, and 
our work builds on these techniques to investigate whether code models exhibit the DKE. Moreover, while prior studies explore confidence misalignment and cognitive bias in general reasoning tasks \cite{singh2024large, wen2024mitigating}, our paper focuses on coding tasks and introduces a formal, domain-specific analysis of the DKE using both absolute and relative confidence metrics.



\section{Methods}

%
%

For our study, we use \emph{multiple-choice questions} $(q, A, a)$ as tasks
where $q$ is the programming-related question, $A$ is the set of answer choices,
and $a$ is the expected answer.
Each question $q$ also belongs to a domain $q \in D$ that is the broad topic
which this question pertains to.
In our setting, the tasks are specific questions about programming and domains
are the individual programming languages.
For example, the question ``Variables of which data types are preceded by
a dollar sign in Perl?'' will have the domain ``Perl''.
For each such task and model $M$, we prompt the model $M$ to answer the question
$q$ given the choices $A$---we say the model $M$ is correct on the task if the
answer $a_M$ produced by $M$ matches $a$.
We define the \emph{actual performance} $\mathsf{AP}(M, D)$  of the model $M$ on a domain
$D$ to the fraction of domain $D$ tasks it is correct on. 

\subsection{Measuring Perceived Performance}

We use two different techniques to measure perceived performance of AI models,
absolute confidence and relative confidence.
For absolute confidence, the model is asked to produce a confidence score in
the range $[0, 1]$ along with its answer.
Model $M$'s absolute confidence $\mathsf{PP}_\mathsf{Abs}(M, D)$ on a domain $D$
is the mean of its absolute confidence scores on individual tasks that belong to
$D$.

%
Previously, relative confidence estimation methods have been shown to produce
more reliable confidence scores than absolute confidence
estimation~\cite{shrivastava2025language}.
%
For every pair of questions $q_i$ and $q_j$, we prompt the model to indicate
which it is more confident in answering.
%
%
These pairwise preferences are aggregated into scalar confidence scores using
two different rank aggregation algorithms, ELO~\cite{elo1978rating} and
TrueSkill~\cite{herbrich2006trueskill}.
%
%
These algorithms treat each question as a ``player'' with $q_i$ ``winning''
against $q_j$ if the model is more confident in answering $q_i$ over $q_j$.
They produce a scalar strength value for each $q_i$ with higher
strengths indicating the model's higher confidence
(see Appendix~\ref{app:relative-confidence}).
We normalize the ELO and TrueSkill scores to the range $[0, 1]$ linearly, and
set the relative confidences $\mathsf{PP}_\mathsf{ELO}(M, D)$ and
$\mathsf{PP}_\mathsf{TrueSkill}(M, D)$ to be the mean strengths of the
questions in $D$.

\begin{figure*}
    \centering
    \includegraphics[width=\linewidth]{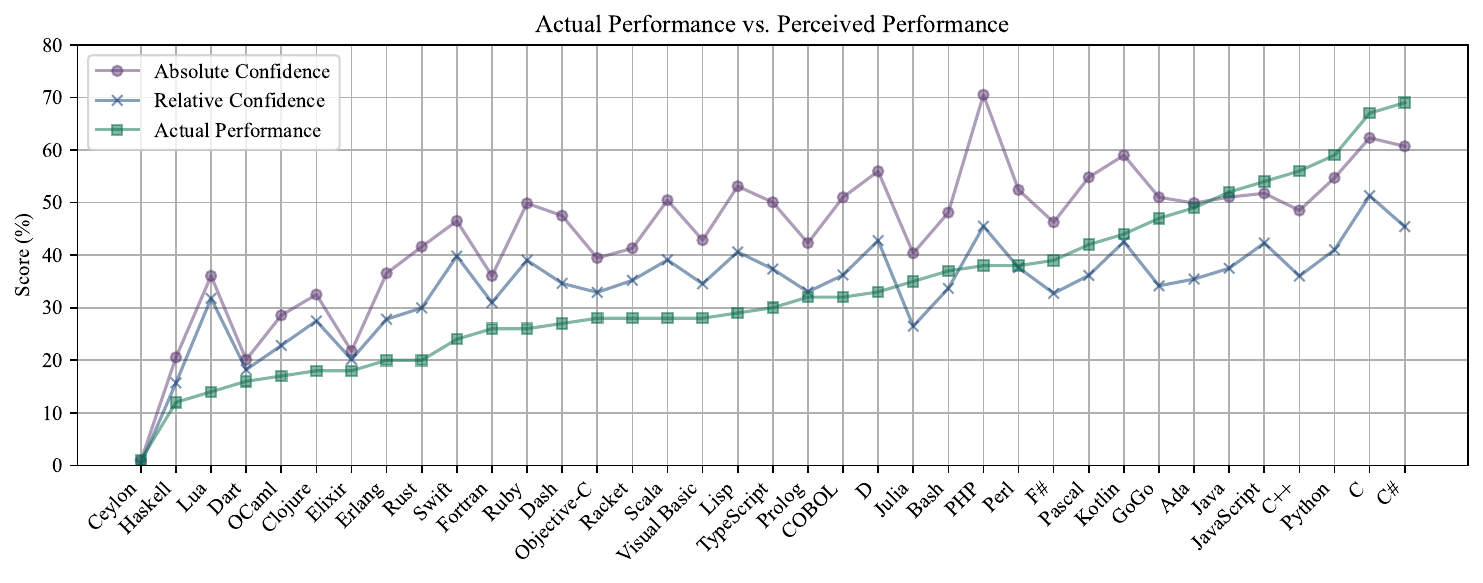}
    \vspace{-4ex}
    \caption{Actual vs. perceived performance for GPT-4o across
    different languages sorted by actual performance}
    \label{fig:intra-dunning-kruger}
    \vspace{-2ex}
\end{figure*}

\begin{figure}
    \centering
    \includegraphics[width=\linewidth]{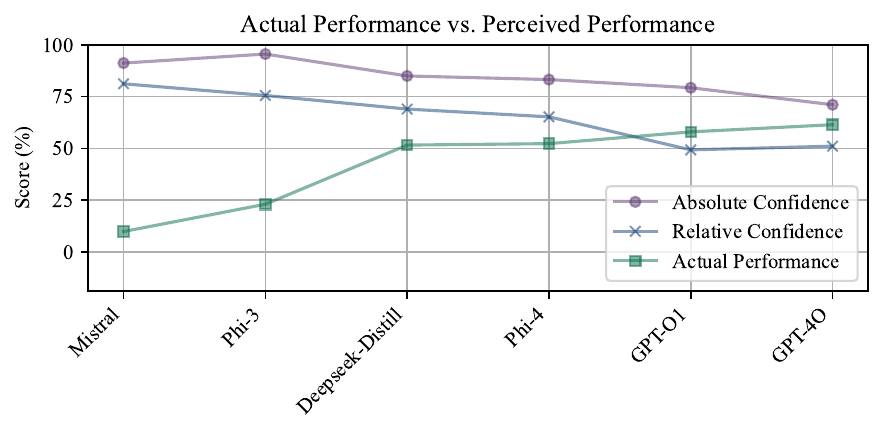}
    \vspace{-5ex}
    \caption{Inter-model DKE}
    \label{fig:inter-dunning-kruger}
    \vspace{-2ex}
\end{figure}

\subsection{Measuring the Dunning-Kruger Effect}

There have been several closely related effects that have all been referred to
under the umbrella term of DKE \cite{kruger1999unskilled}.
Here, we consider two specific variants from the literature---the
intra-participant \cite{muthukrishna2018overconfidence, moore2018overconfidence} and inter-participant versions \cite{dunning2003people, hodges2001difficulties, edwards2003medical, haun2000assessing}.
In the intra-participant version, the question is ``Does a single participant
over-estimate their performance more in domains where they have low actual
performance?'' and for the inter-participant version, it is ``Do participants
who show low actual performance over-estimate their performance more?''
%
%

%
For the intra-participant version, we fix $M$ and measure the
\emph{over-confidence} per domain $D$:
\[
\Delta_{\text{overconf}}(M, D) = \mathsf{PP}(M, D) - \mathsf{AP}(M, D) 
\]
where $\mathsf{PP}$ is one of $\mathsf{PP}_\mathsf{Abs}$,
$\mathsf{PP}_\mathsf{ELO}$, or $\mathsf{PP}_\mathsf{TrueSkill}$.
For the inter-participant version, we have: 
\[
\Delta_{\text{overconf}}(M) = \mathbb{E}_D[\mathsf{PP}(M, D)] - \mathbb{E}_D[\mathsf{AP}(M, D)].
\]
Higher $\Delta_{\text{overconf}}$ in regimes with low actual
performance is indicative of the corresponding DKE.

\ignore{
\arsays{Wrote some very rough text up to here}

To assess the Dunning-Kruger effect, we compare the model’s confidence scores $C(x)$ (from either method) with its actual performance $R(\hat{y}, y)$. We stratify questions by difficulty and model performance, and analyze whether lower-performing models or domains exhibit inflated confidence. Overconfidence is quantified as the difference between average confidence and accuracy:

\[
\Delta_{\text{overconf}} = \mathbb{E}[C(x)] - \mathbb{E}[R(\hat{y}, y)]
\]

A positive $\Delta_{\text{overconf}}$ in low-competence regimes is indicative of the Dunning-Kruger effect. 
}

\section{Results}


We evaluate the presence of the Dunning-Kruger Effect (DKE) in six
large language models (LLMs) across 37 programming languages using
multiple-choice question answering (MCQA) tasks.
The multiple-choice QA data is derived from publicly-available data
called CodeNet \cite{puri2021codenet}. More details on the implementation and
experimental setup is in Appendix~\ref{sec:appendix-implementation-details}

\subsection{Do code models exhibit the DKE?}

\ignore{
We consider two specific variants of DKE from the literature - the inter-participant version and the intra-participant version. The inter-participant version translates into the inter-model interpretation of DKE wherein we compare the performance across models of varying capabilities. The intra-participant version translates to the intra-model or inter-domain interpretation of DKE wherein we examine the performance of a model across different domains.
}

\paragraph{Inter-model interpretation of DKE}

In the inter-model analysis, we observe the DKE pattern: lower-performing
models consistently overestimate their capabilities, while higher-performing
models exhibit more calibrated or even underconfident behavior. As shown by
Fig.~\ref{fig:inter-dunning-kruger}, models such as Mistral and Phi-3 display
a gap between perceived and actual performance. In contrast, models like GPT-4O demonstrate more alignment
between perceived and actual performance, especially in relative
confidence estimates. Interestingly, the relative confidence curve intersects with
the actual performance curve, suggesting that
higher-performing models may become under-confident, an effect not
captured by absolute confidence alone.

\paragraph{Intra-domain interpretation of DKE}

\begin{wraptable}{r}{0.6\textwidth}
\small
\centering
\caption{Correlation between overestimation (AC - RC) and true performance for different model setups as participants.}
\label{tab:personas}
\begin{tabular}{lrrr}
\toprule
\textbf{\textit{Setup}} & \textbf{\textit{Spearman}} & \textbf{\textit{Pearson}} & \textbf{\textit{Kendall}}\\
\midrule
Different Models & 0.775 & 0.640 & 0.592 \\
Different Personas & 0.712 & 0.618 & 0.587 \\
Diversity Sampling & 0.821 & 0.670 & 0.611 \\
Prompt Phrasing & 0.750 & 0.633 & 0.581 \\
\bottomrule
\end{tabular}
\end{wraptable}

The intra-model analysis further supports the presence of DKE. Figure~\ref{fig:intra-dunning-kruger} presents model performance across different domains (programming languages), ordered by actual performance. In domains where models perform poorly, typically rare or low-resource languages such as COBOL, Prolog, and Ceylon, we observe higher overconfidence. Conversely, in high-performing domains like Python and JavaScript, models tend to be better calibrated or even underconfident. This domain-level overestimation is consistent across both absolute and relative confidence measures, reinforcing the hypothesis that models are less aware of their limitations in unfamiliar domains.

\begin{wraptable}{r}{0.6\textwidth}
\small
\centering
\caption{Correlation Between Overestimation (AC - RC) and True Performance for Domains and Models}
\label{tab:ac_rc_comparison}
\begin{tabular}{ll l l}
\toprule
\textbf{\textit{Category}} & \textbf{\textit{Method}} & {\textbf{\textit{Corr $\rho$ / Tau $\tau$}}} & \textbf{\textit{p value}} \\
\midrule
\multirow{3}{*}{Domains}
  & Spearman & 0.775 & $1.797 \times 10^{-8}$ \\
  & Pearson  & 0.640 & $2.019 \times 10^{-5}$ \\
  & Kendall  & 0.592 & $3.058 \times 10^{-7}$ \\
\midrule
\multirow{3}{*}{Models}
  & Spearman & 0.775 & $1.797 \times 10^{-8}$ \\
  & Pearson  & 0.640 & $2.019 \times 10^{-5}$ \\
  & Kendall  & 0.592 & $3.058 \times 10^{-7}$ \\
\bottomrule
\end{tabular}
\vspace{-2ex}
\end{wraptable}

Treating models as participants introduces certain biases: (1) the limited number of models restricts diversity; (2) intra-model variability is overlooked; and (3) the underlying distribution may differ.
To address these concerns, we explored three alternative evaluation setups:
(1) Assigning distinct personas to models 
(2) Sampling model responses at a higher temperature ($T=0.75$) to simulate varied participant answers;
(3) Generating multiple paraphrased versions of the prompt and considering each as a separate participant, with 10 paraphrases created (see appendix for the prompt template \ref{sec:relative-confidence-prompt}).
Table~\ref{tab:personas} summarizes the DKE strength observed across these setups. Notably, DKE is consistently present in all configurations and is most pronounced when sampling multiple responses from the same model at high temperature.

\subsection{Analysis of Perceived Performance}

\paragraph{Absolute Confidence vs. Relative Confidence}
To quantify these trends, we compute the correlation between overestimation (perceived minus actual performance) and true performance across both models and domains. Table~\ref{tab:ac_rc_comparison} includes the correlation between (a) actual performance across domains vs. overestimation of performance (AC - RC) and (b) actual performance across models vs. overestimation of perceived performance (AC - RC). The results suggest that the overestimation of perceived performance is higher for models and domains that are more high performing. This indicates that AC becomes an unreliable measure of perceived performance, especially as we encounter increasingly better performing models or domains where LLMs achieve higher performance.

\paragraph{DKE on Specialized Models}

\begin{wraptable}{r}{0.55\textwidth}
\small
\centering
\caption{Correlation Between Overestimation (AC - RC) and True Performance for single domain specialized vs multiple models.}
\label{tab:model_specialization}
\begin{tabular}{ll l}
\toprule
\textbf{\textit{Model}} & \textbf{\textit{Metric}} & {\textbf{\textit{Corr $\rho$ / Tau $\tau$}}} \\
\midrule
\multirow{3}{*}{Base}
  & Spearman & 0.775 \\
  & Pearson  & 0.640 \\
  & Kendall  & 0.592 \\
\midrule
\multirow{3}{*}{Single Specialized}
  & Spearman & 0.921 \\
  & Pearson  & 0.883 \\
  & Kendall  & 0.734 \\
\midrule
\multirow{3}{*}{Multi Specialized}
  & Spearman & 0.831 \\
  & Pearson  & 0.755 \\
  & Kendall  & 0.676 \\
\bottomrule
\end{tabular}
\vspace{-2ex}
\end{wraptable}

Specialized models may exhibit different calibration dynamics due to narrower training distributions or domain-specific optimization. In particular, we believe that contrasting domain-specialized and generalist models could help disentangle whether DKE-like effects arise from general cognitive miscalibration or from mismatches between training exposure and task domain.

To evaluate this, we measure DKE for models in three settings: (1) base ssetup, (2) trained on a single domain, (3) trained on multiple domain. We use the MultiPL-E dataset \cite{multiple} for this and consider 8 languages (Ada, Dart, Prolog, Swift, C++, Python, C\#, Elixir) where each language is a domain. For single domain training we pick one language and train on that while for multi domain we train on all 8 languages.

Table~\ref{tab:model_specialization} shows the correlation for base, single domain specialized and multi domain specialized models. We see that specialization increased the strength of DKE. Furthermore, single domain specialization shows stronger DKE compared to multi domain highlighting that DKE scales with the degree of specialization.

\vspace{-2ex}

\paragraph{Impact of Rarity of Programming Language}

\begin{wraptable}{r}{0.6\textwidth}
\small
\centering
\caption{Perceived performance vs. Rarity Ranking}
\label{tab:correlation}
\begin{tabular}{llcc}
\toprule
\textbf{\textit{Ranking}} & \textbf{\textit{Method}} & {\textbf{\textit{Corr $\rho$ / Tau $\tau$}}} & {\textbf{\textit{p} ($10^{-3}$)}} \\
\midrule
\multirow{2}{*}{GitHub} 
  & \textit{Spearman} & 0.797 & $1.318$ \\
  & \textit{Kendall}  & 0.690 & $3.935$ \\
\midrule
\multirow{2}{*}{IEEE} 
  & \textit{Spearman} & 0.683 & $5.863$ \\
  & \textit{Kendall}  & 0.529 & $8.970$ \\
\midrule
\multirow{2}{*}{TIOBE} 
  & \textit{Spearman} & 0.741 & $0.234$ \\
  & \textit{Kendall}  & 0.662 & $0.354$ \\
\bottomrule
\end{tabular}
\end{wraptable}

We also investigate the relationship between domain rarity and overconfidence.
Table~\ref{tab:correlation} shows the correlation of perceived performance with
(a) GitHub ranking (most used languages on GitHub)
\cite{GitHubRanking}, (b) IEEE popularity ranking \cite{IEEERanking}, and (c)
TIOBE index \cite{TIOBERanking}. 
Across all
three sources, we observe a consistent trend: models exhibit higher
overconfidence in rarer languages. For instance, GitHub ranking shows a correlation of 0.797 with perceived confidence, highlighting that rarity
is a predictor of overconfidence.

\vspace{-1ex}

\begin{wraptable}{r}{0.6\textwidth}
\small
\centering
\caption{DKE measurement on code generation tasks.}
\label{tab:code-gen-results}
\begin{tabular}{lr}
\toprule
\textbf{\textit{Metric}} & {\textbf{\textit{Corr $\rho$ / Tau $\tau$}}}\\
\midrule
Spearman's $\rho$ & 0.734\\
Pearson's $\rho$ & 0.625\\
Kendall's $\tau$ & 0.580\\
\bottomrule
\end{tabular}
\end{wraptable}

\vspace{-2ex}

\paragraph{Non-MCQ style tasks}
Focusing exclusively on MCQ programming questions limits the generalizability of the findings to more open-ended tasks such as code generation or broader NLP applications. Our choice of multiple-choice format was intentional: it allows for precise measurement of correctness and confidence, which is essential for quantifying miscalibration and identifying Dunning–Kruger-like patterns. That said, real-world tasks often involve partial correctness, ambiguity, or creative reasoning, where confidence calibration may manifest differently. Table~\ref{tab:code-gen-results} shows the DKE correlation on code generation dataset (MultiPL-E \cite{multiple}). We consider 8 languages--(Ada, Dart, Prolog, Swift, C++, Python, C\#, Elixir). We see that the effect can be observed in code generation task too but is much weaker. This could be due to the challenge in accurately measuring the confidence in the task.

\section{Discussion and Conclusion}


\paragraph{Reviewer agents.}
Reviewer agents are a common and successful design-pattern in multi-agent
systems~\cite{li2024llms, zhou2025multi, gu2024survey, jin2024agentreview}.
%
%
While not all reviewer agents directly align with self-evaluation, our findings highlight the need to further investigate what auxiliary information improves their reliability.

\vspace{-2ex}

\paragraph{Cognitive Effect or Statistical Effect?}
%

There has been significant debate in the psychology and cognitive science
community on whether DKE is a ``real'' effect with an underlying cognitive
cause, or if it is ``merely'' a statistical effect akin to regression to the mean~\cite{magnus2022statistical}.
Our findings indicate that AI models exhibit DKE-like behavior, raising three possibilities: (a) DKE arises from the same cognitive mechanism in humans and AI models; (b) DKE is cognitive, but mechanisms differ; or (c) DKE is purely statistical. Our results prompt further investigation into options (a) and (b), as determining the origin and scope of DKE in AI models and its similarity to human cognition requires substantial future research.

\vspace{-2ex}

\paragraph{Shared Authorship with LLMs.}

As AI models become central to creative and technical workflows, authorship is increasingly collaborative. Our results show that LLMs, like human partners, may misjudge their competence, especially in unfamiliar domains. This necessitates new frameworks for transparent self-assessment and mutual trust in co-creation. Importantly, overconfidence in AI models stems from technical factors such as training data or architecture, not self-awareness or intent. Thus, while AI may mirror human cognitive patterns, its underlying mechanisms remain fundamentally distinct.

\vspace{-2ex}

\paragraph{Conclusion.}
Our study is an initial foray into studying whether AI models display cognitive
biases that have been previously observed in humans.
Our results show that AI models, specifically in the context of answering
programming related questions, display DKE-like behaviour.
%
This points to a rich set of future research directions related to
the strength and scope of the DKE in models, as well as other self-assessment
related cognitive biases including the hard-easy effect or IOED~\cite{juslin1993explanation, levin2000change, chromik2021think}.


\section{Limitations}

\paragraph{Domain and task choices.}
%

This study is limited to the programming domain, so results may not generalize to other areas where AI language models are used. We focus on multiple-choice question answering, which simplifies performance estimation by avoiding issues like partial correctness or varied response styles. For broader conclusions about DKE in AI models, future work should expand both the domains and task types considered.

\vspace{-2ex}

\paragraph{Measurements.}
A key measurement in our study is perceived performance, i.e., the model's confidence in its answers. Prior work has highlighted limitations in models' ability to assign reliable confidence scores~\cite{shorinwa2024survey, shen2024thermometer, yang2024can, li2024showing}, with relative confidence~\cite{shrivastava2025language} suggested as an alternative. Assessing perceived performance in AI models is less straightforward than in humans, adding complexity to the study of self-assessment biases and posing a threat to the validity of our results.
%
%
\vspace{-2ex}

\paragraph{Explanations for the effect.}
Many explanations have been hypothesized to be the underlying cause of the DKE
in humans~\cite{ehrlinger2008unskilled}.
Our study intentionally does not attempt to compare or contrast the underlying
explanation of the DKE in humans and AI models.
Many explanations attributed to the DKE in humans are not directly applicable to
AI models.
Human DKE explanations rely on causes such as overly positive prior
beliefs~\cite{ehrlinger2008unskilled}, the distribution of over- and under-performers in the
human population, or lack of incentive for accurate self-assessments---none of
these apply directly to AI models.
One potential explanation that may be common to both humans and AI models is the
meta-cognitive explanation, which states that assessing the quality of a
performance of a skill is a crucial part of acquiring a skill.
This explanation can potentially be tested experimentally in AI models with a
controlled study of different training strategies and whether they all lead to
simultaneous improvements in performance and in the ability to assess quality of
performance.
However, this study is significantly beyond the scope of this paper, and we
leave it for future work.

\ignore{
%
\begin{itemize}
    \item Restricted domain, restricted type of tasks
    \item Confidence measurements (not as accurate as for humans)
    \item We don't do any attempt on explanations
    \item Rarity is measured by popularity scores (for TIOBE and IEEE), but
        other papers use it.
    \item ???
\end{itemize}
}

\bibliographystyle{unsrtnat}
\bibliography{custom}

\clearpage
\appendix

\section{Measuring Relative Confidence}
\label{app:relative-confidence}

\ignore{
\subsection{Absolute Confidence Estimation}

In the absolute setting, the model is prompted to answer a programming question and simultaneously provide a scalar confidence score in the range $[0, 1]$. Formally, for a given input $x$, let $\hat{y}(x)$ denote the model’s predicted answer and $y(x)$ the ground-truth answer. The model’s absolute confidence is denoted as:

\begin{equation*}
    C_{\text{abs}}(x, \hat{y}) \in [0, 1]
\end{equation*}
The correctness of the prediction is defined as:

\begin{equation*}
R(\hat{y}, y) = 
\begin{cases}
1 & \text{if } \hat{y}(x) = y(x) \\
0 & \text{otherwise}
\end{cases}
\end{equation*}
}

Here, we elaborate on the algorithms we use to convert pairwise confidence
preferences into scalar relative confidence scores for each question.
For every pair of questions $q_i$ and $q_j$, we prompt the model to indicate
which it is more confident in answering to produce a set $P$ of pairwise
preferences $q_i < q_j$.
Below, we present the details of the ELO and TrueSkill methods to convert these
preferences to scalar confidence scores.

\subsection{Confidence Estimation Using Elo Rating}

We treat each question as a ``player'' in a tournament, where each pairwise preference is interpreted as a match outcome. All questions are initialized with the same Elo score \cite{elo1978rating} (in our case, 1000). For each preference pair $q_i < q_j \in P$, where \( q_i \) is the preferred (winning) question and \( q_j \) is the less preferred (losing) question, we compute the expected win probability using the logistic function:

\[
P(i \text{ wins}) = \frac{1}{1 + 10^{(S_j - S_i)/K}}
\]

where \( S_i \) and \( S_j \) are the Elo scores of questions \( q_i \) and \( q_j \), and \( K \) is a sensitivity factor. The scores are then updated as follows:

\[
S_i \leftarrow S_i + K \cdot (1 - P(i \text{ wins}))
\]
\[
S_j \leftarrow S_j - K \cdot P(i \text{ wins})
\]

This update process is repeated for all preference pairs over multiple iterations to allow scores to converge. The final Elo scores are normalized using min-max scaling to the range \([0, 100]\) to produce interpretable confidence scores:

\[
\text{Confidence}(q_i) = \frac{S_i - \min(S)}{\max(S) - \min(S)}
\]

\subsection{Confidence Estimation Using TrueSkill}

As an alternative to Elo, we also implement confidence estimation using the TrueSkill rating system \cite{herbrich2006trueskill}, which models each question’s confidence as a Gaussian distribution over skill: \( \mathcal{N}(\mu, \sigma^2) \), where \( \mu \) represents the estimated confidence and \( \sigma \) the uncertainty. 

For each preference pair \( q_i < q_j \in P \), where \( q_i \) is preferred over \( q_j \), we update the distributions of both questions using Bayesian inference. The update is performed using the TrueSkill factor graph model, which adjusts both \( \mu \) and \( \sigma \) based on the observed outcome and the prior distributions. After processing all preference pairs, we extract the mean \( \mu_i \) of each question’s distribution as its raw confidence score. These scores are then min-max normalized to the range \([0, 100]\) as in the Elo method.

This methodology enables robust and interpretable confidence estimation by
leveraging the model’s relative preferences, rather than relying on coarse,
absolute confidence scores.

\subsection{Dataset}
\label{sec:appendix-dataset}

We create the MCQA problems for the codenet tasks \cite{puri2021codenet} covering (1) code generation; (2) code understanding; (3) code syntax; and (4) code repair. For doing this scalably we 
Table~\ref{tab:datasets} summarizes the number of tasks used for each domain.

\begin{table}[htbp]
\small
\centering
\caption{Data Statistics}
\label{tab:datasets}
\begin{tabular}{lr}
\toprule
\textbf{\textit{Domain}} & \textbf{\textit{Number of Samples}}\\
\midrule
Ada & 118\\
Bash & 1000\\
C & 897\\
C\# & 1000\\
C++ & 1000\\
COBOL & 1000\\
Ceylon & 90\\
Clojure & 430\\
D & 1000\\
Dart & 195\\
Dash & 155\\
Elixir & 205\\
Erland & 141\\
F\# & 1000\\
Fortran & 1000\\
Go & 1000\\
Haskell & 1000\\
Java & 1000\\
JavaScript & 1000\\
Julia & 1000\\
Lisp & 1000\\
Kotlin & 1000\\
Lua & 1000\\
OCaml & 1000\\
Objective-C & 727 \\
PHP & 1000\\
Pascal & 1000\\
Perl & 1000\\
Prolog & 231\\
Python & 1000\\
Racket & 145\\
Ruby & 1000\\
Rust & 1000\\
Scala & 1000\\
Swift & 1000\\
TypeScript & 1000\\
Visual Basic & 987\\
\bottomrule
\end{tabular}
\end{table}

\begin{figure*}[htbp]
  \centering

  \begin{subfigure}[b]{0.3\textwidth}
    \includegraphics[width=\textwidth]{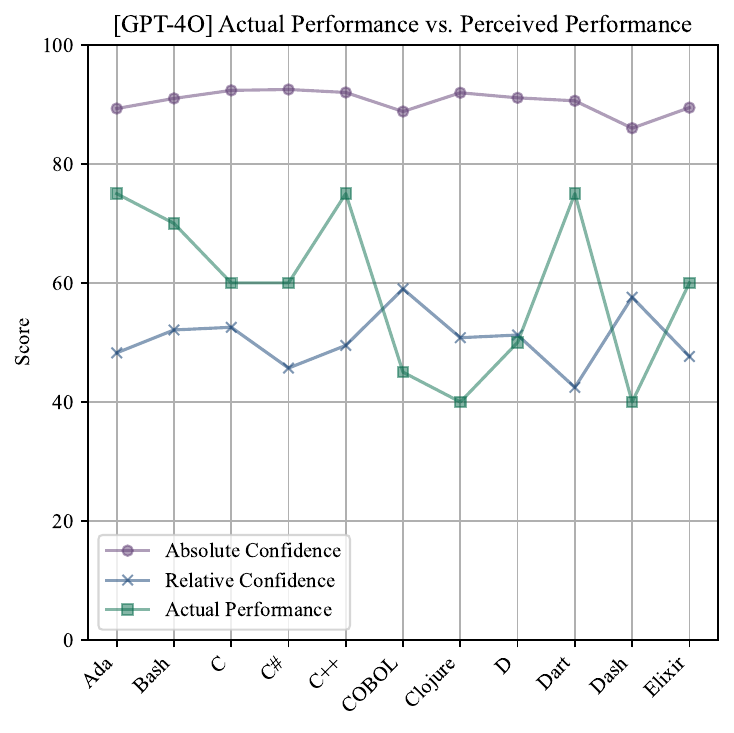}
    \caption{GPT-4O}
  \end{subfigure}
  \hfill
  \begin{subfigure}[b]{0.3\textwidth}
    \includegraphics[width=\textwidth]{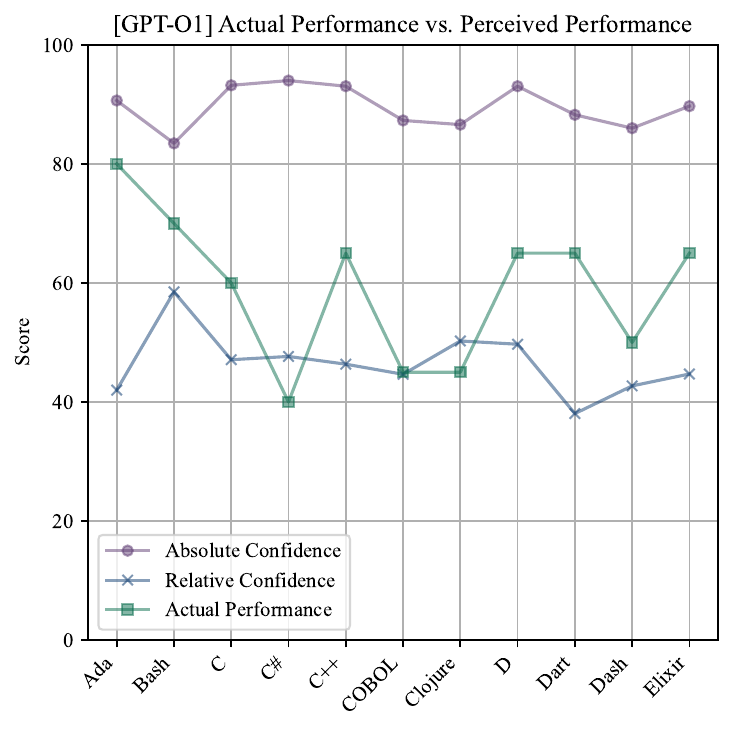}
    \caption{GPT-O1}
  \end{subfigure}
  \hfill
  \begin{subfigure}[b]{0.3\textwidth}
    \includegraphics[width=\textwidth]{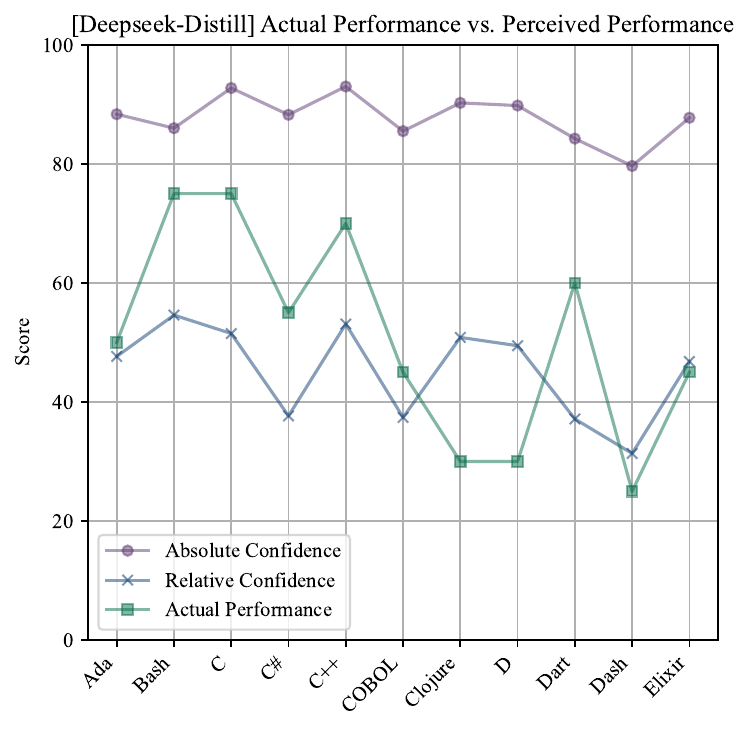}
    \caption{Deepseek-Distill}
  \end{subfigure}

  \vspace{1em}

  \begin{subfigure}[b]{0.3\textwidth}
    \includegraphics[width=\textwidth]{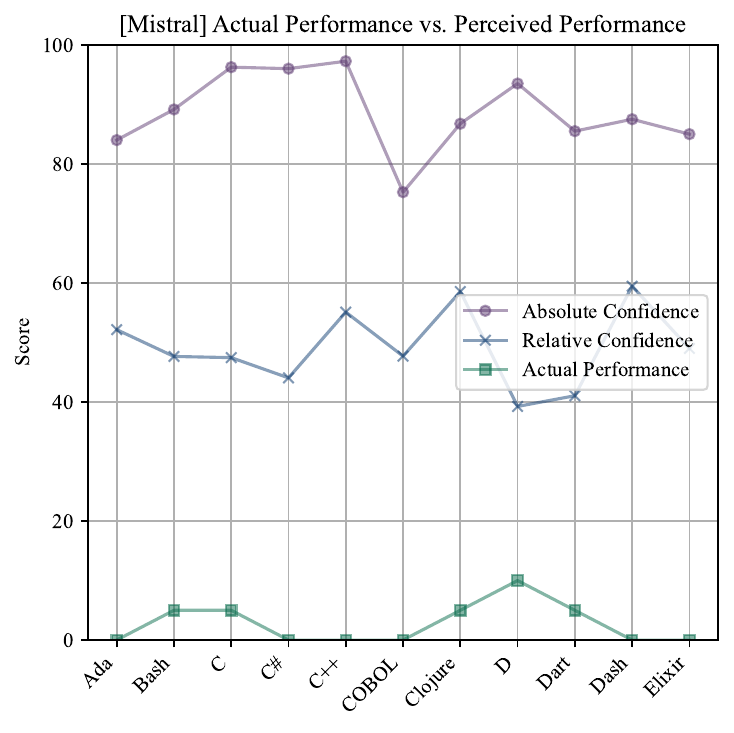}
    \caption{Mistral}
  \end{subfigure}
  \hfill
  \begin{subfigure}[b]{0.3\textwidth}
    \includegraphics[width=\textwidth]{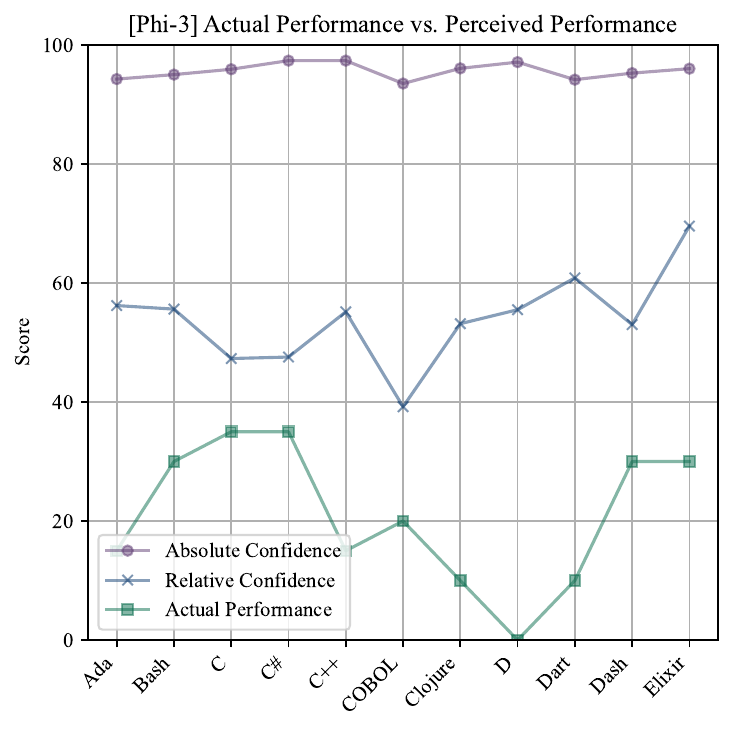}
    \caption{Phi-3}
  \end{subfigure}
  \hfill
  \begin{subfigure}[b]{0.3\textwidth}
    \includegraphics[width=\textwidth]{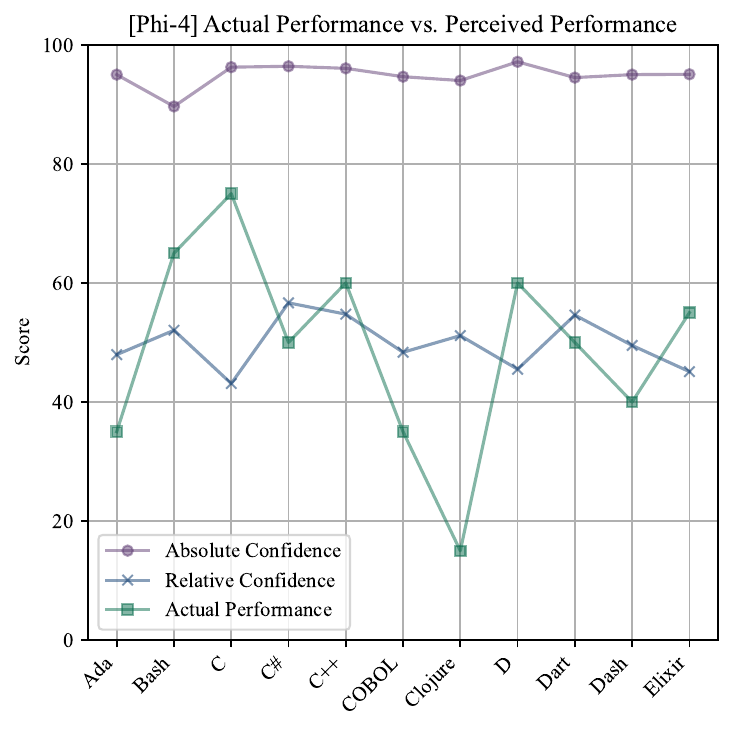}
    \caption{Phi-4}
  \end{subfigure}

  \caption{Dunning-Kruger plots for various models.}
  \label{app:fig:vary_model}
\end{figure*}

\subsection{Implementation Details}
\label{sec:appendix-implementation-details}

The programming languages include in the study are - Ada, Bash, C, C\#, C++, COBOL, Ceylon, Clojure, D, Dart, Dash, Elixir, Erland, F\#, Fortran, Go, Haskell, Java, JavaScript, Julia, Lisp, Kotlin, Lua, OCaml, Objective-C, PHP, Pascal, Perl, Prolog, Python, Racket, Ruby, Rust, Scala, Swift, TypeScript and Visual Basic. To generate the pairwise question preference data, we randomly sample 5 questions to generate multiple comparisons per question. The model's preferences are parsed to construct a directed graph of confidence judgements wherein each comparison yields a winner-loser pair, forming the basis for confidence ranking. For Elo rating ranking estimation, ratings are initialized randomly at 1000 for each question and updated iteratively based on outcome of each comparison. The final scores are normalized to a 0--100 scale. The win probability is scaled using a sensitivity factor, $K$ which is set at 400 following hyperparameters selected in previous work \cite{shrivastava2025language}. The win probabilities are also estimated over 10 repetitions to allow scores to converge. In the TrueSkill ranking system, the questions are initialized with default values \( \mu = 25.0 \) and \( \sigma = 8.333 \), following standard TrueSkill settings implemented using the Python package. Similar to the Elo rating method, the rankings in the TrueSkill method are normalized to 0-100 scale.

\paragraph{Models and Sizes}
We use GPT-4O (size unknown), GPT-O1 (size unknown), Deepseek-Distill (70B), Mistral (7B), Phi-3 (8B) and Phi-4 (20B) for this paper.

\subsection{Inter-domain results for different models}
Figure~\ref{app:fig:vary_model} shows the individual plots for inter-model DKE for different domains. We see that the effect can be seen for all models across varying domains. For very small models like Mistral and Phi-3 (less than 8B) we see that the effect is less apparent as the models overall performance is very low and they generally overestimate their performance.

\subsection{Inter-model results for different domains}
\begin{figure*}[htbp]
  \centering

  \begin{subfigure}[b]{0.3\textwidth}
    \includegraphics[width=\textwidth]{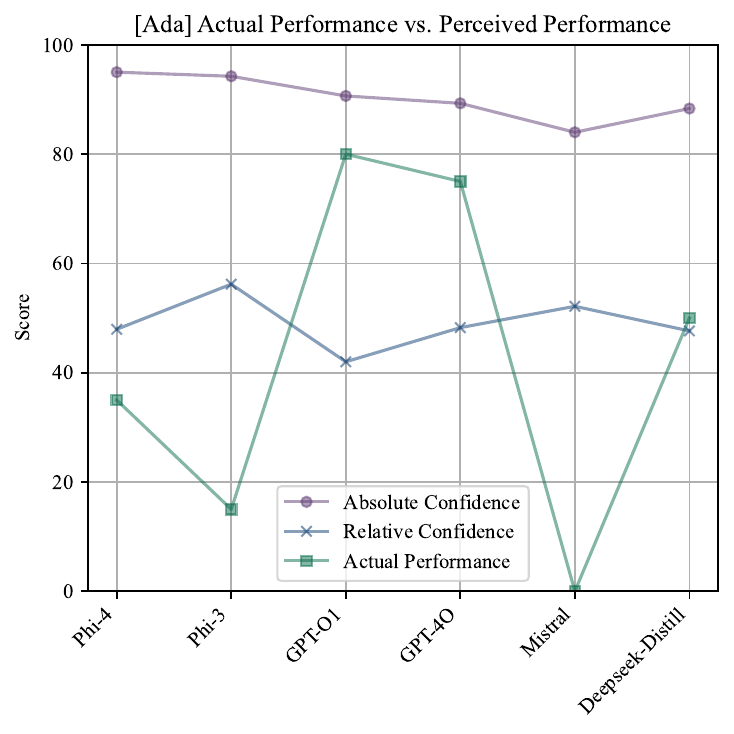}
    \caption{Ada}
  \end{subfigure}
  \hfill
  \begin{subfigure}[b]{0.3\textwidth}
    \includegraphics[width=\textwidth]{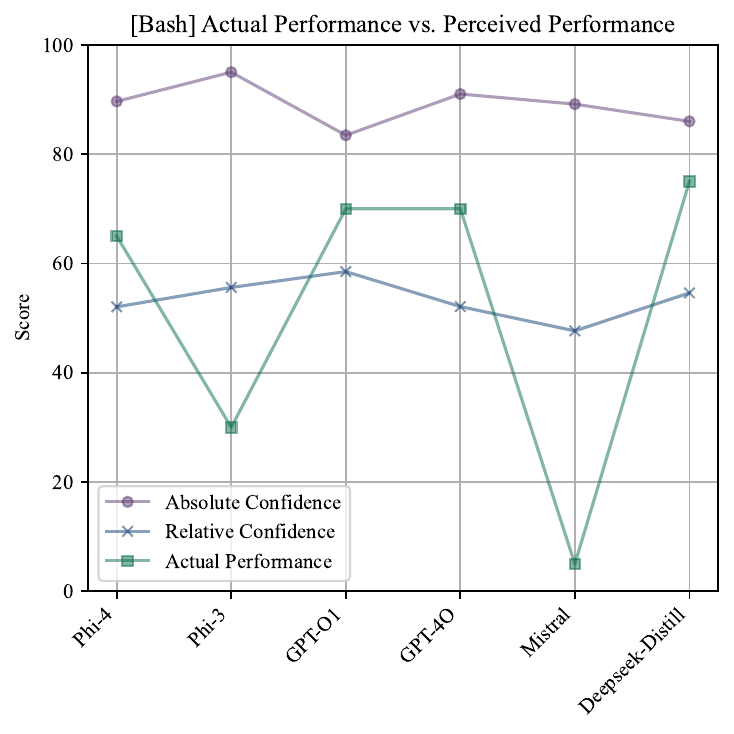}
    \caption{Bash}
  \end{subfigure}
  \hfill
  \begin{subfigure}[b]{0.3\textwidth}
    \includegraphics[width=\textwidth]{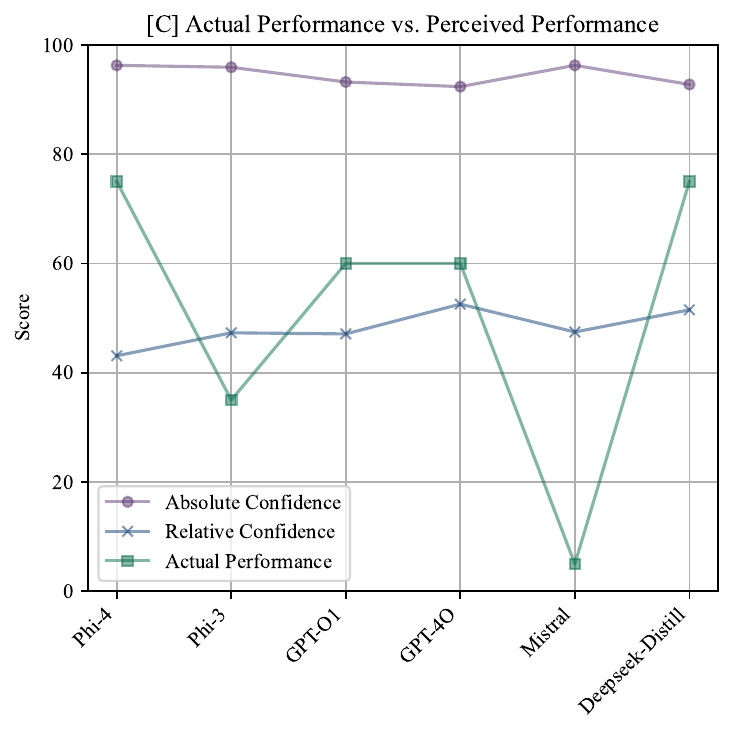}
    \caption{C}
  \end{subfigure}

  \vspace{1em}

  \begin{subfigure}[b]{0.3\textwidth}
    \includegraphics[width=\textwidth]{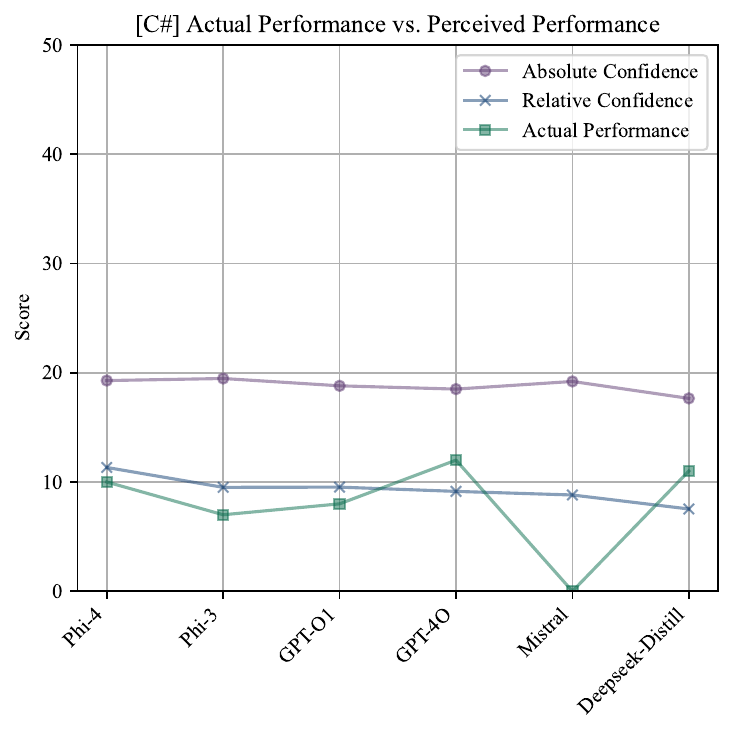}
    \caption{C\#}
  \end{subfigure}
  \hfill
  \begin{subfigure}[b]{0.3\textwidth}
    \includegraphics[width=\textwidth]{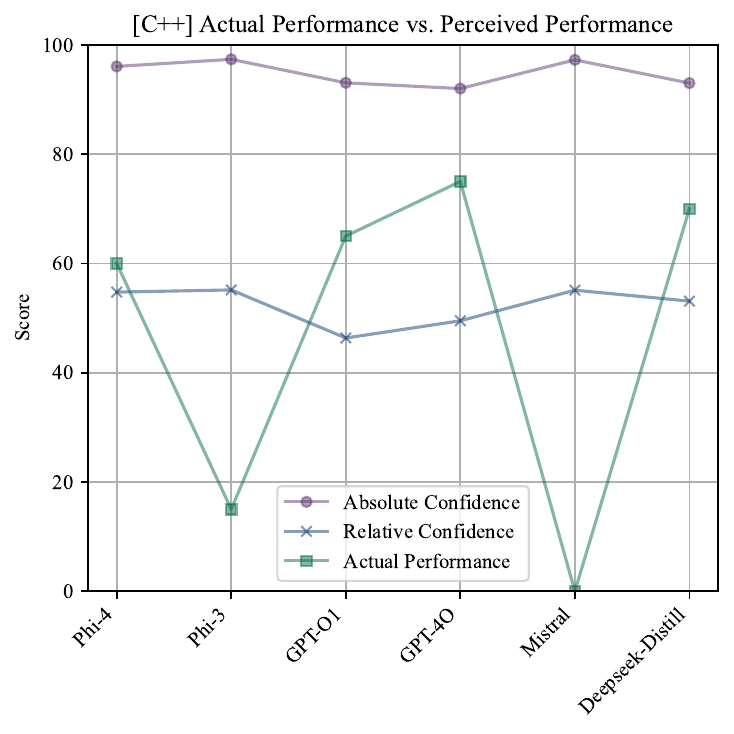}
    \caption{C++}
  \end{subfigure}
  \hfill
  \begin{subfigure}[b]{0.3\textwidth}
    \includegraphics[width=\textwidth]{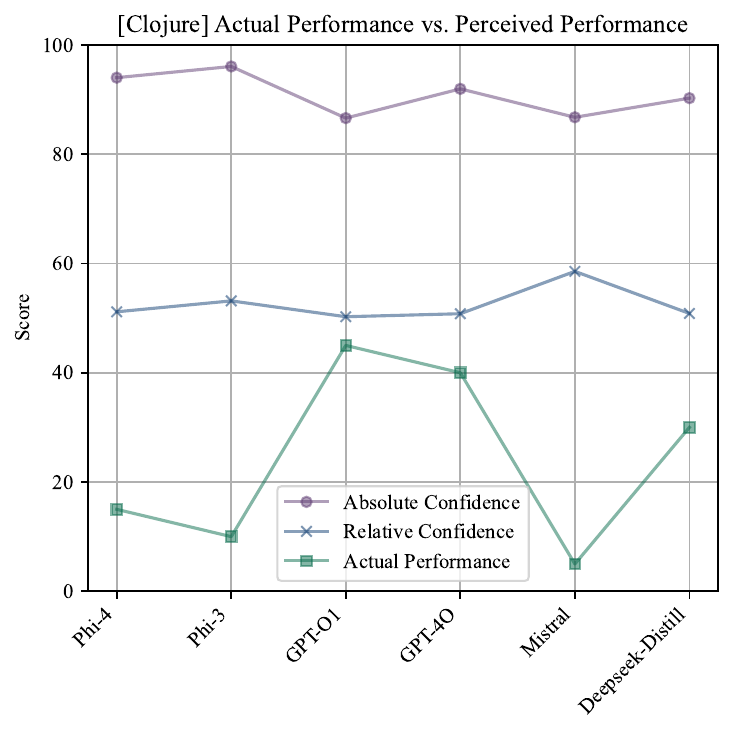}
    \caption{Clojure}
  \end{subfigure}

  \vspace{1em}

  \begin{subfigure}[b]{0.3\textwidth}
    \includegraphics[width=\textwidth]{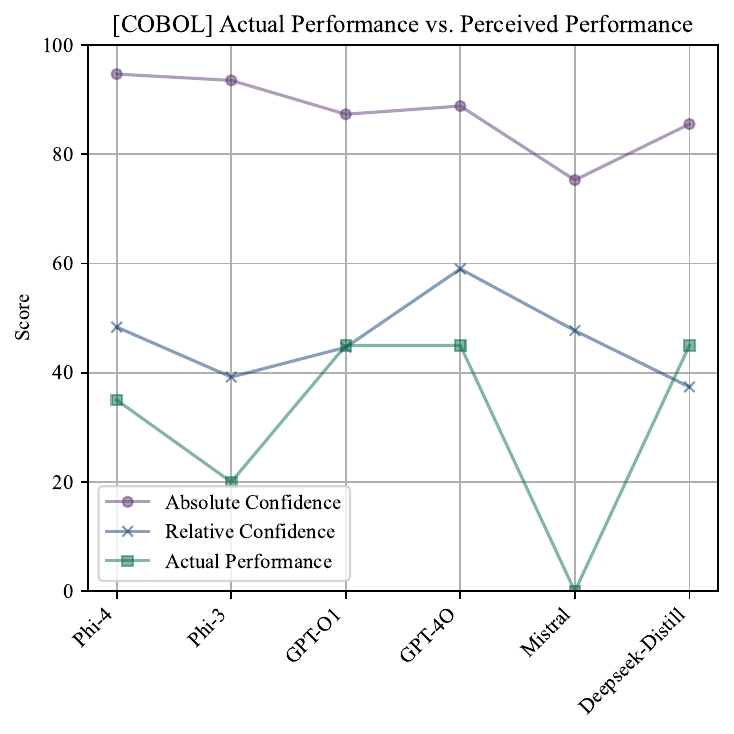}
    \caption{COBOL}
  \end{subfigure}
  \hfill
  \begin{subfigure}[b]{0.3\textwidth}
    \includegraphics[width=\textwidth]{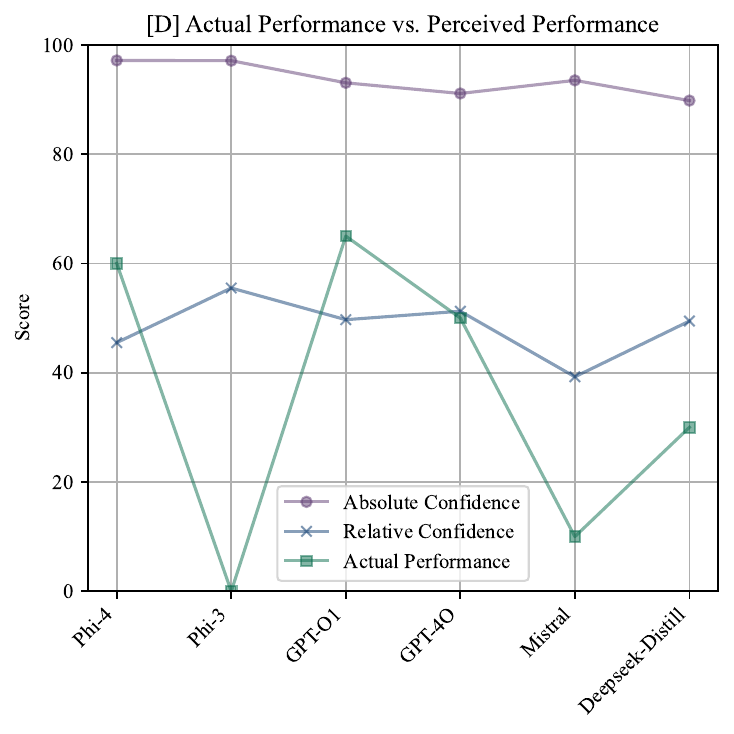}
    \caption{D}
  \end{subfigure}
  \hfill
  \begin{subfigure}[b]{0.3\textwidth}
    \includegraphics[width=\textwidth]{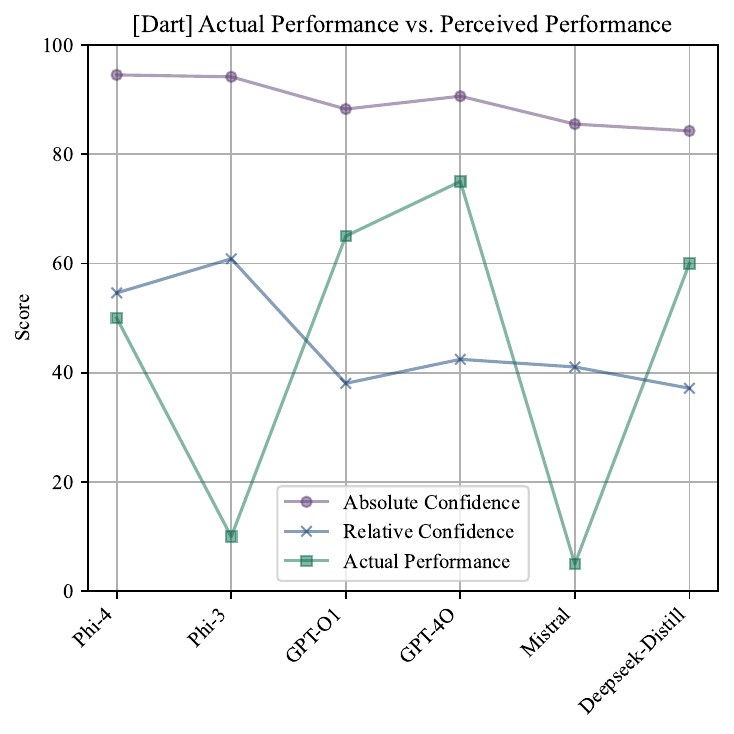}
    \caption{Dart}
  \end{subfigure}

  \vspace{1em}

  \hspace*{\fill}
  \begin{subfigure}[b]{0.3\textwidth}
    \includegraphics[width=\textwidth]{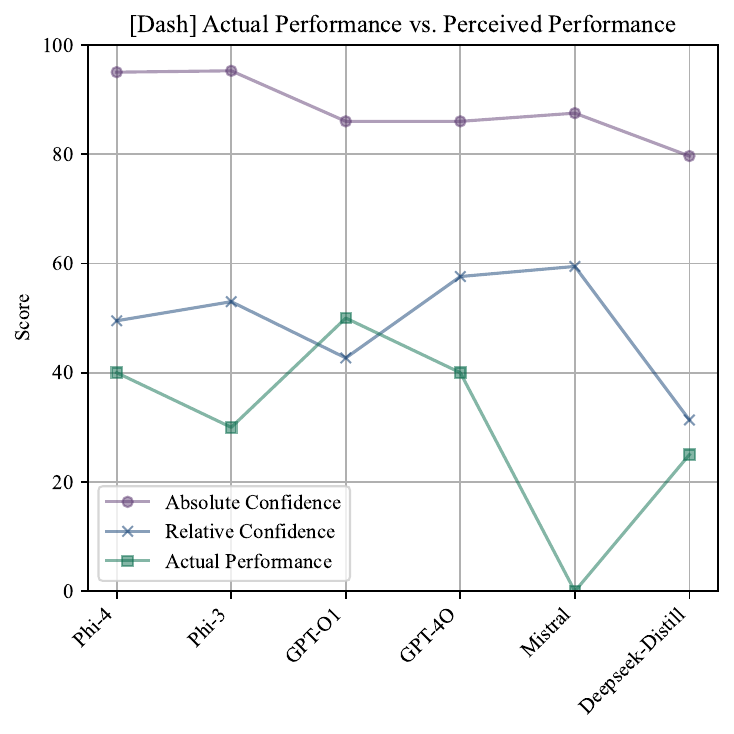}
    \caption{Dash}
  \end{subfigure}
  \hfill
  \begin{subfigure}[b]{0.3\textwidth}
    \includegraphics[width=\textwidth]{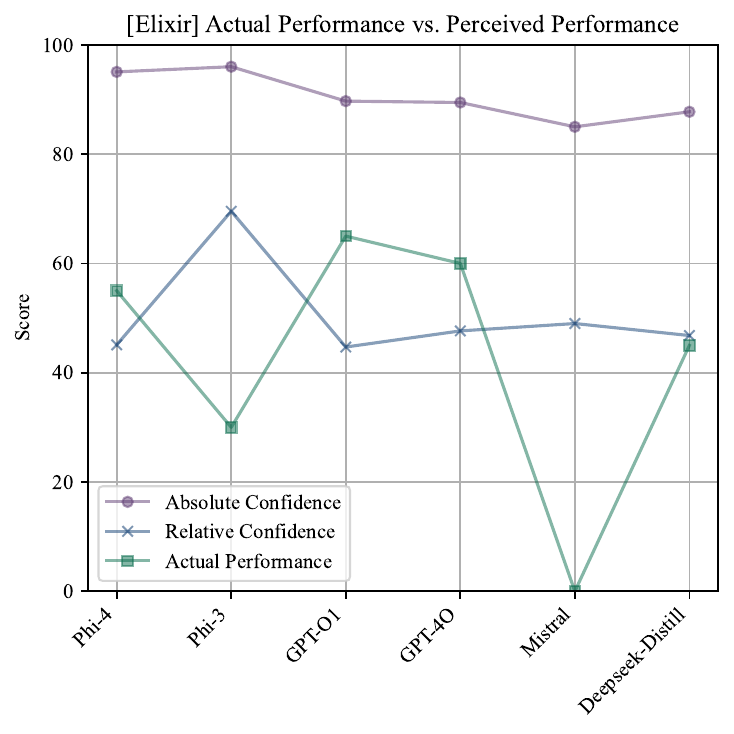}
    \caption{Elixir}
  \end{subfigure}
  \hspace*{\fill}

  \caption{Dunning-Kruger plots for various programming languages.}
  \label{app:fig:vary_domain}
\end{figure*}

Figure~\ref{app:fig:vary_domain} shows the individual plots for inter-model DKE for different domains. We see that the effect can be seen for all domains across varying models.

\subsection{Prompts}
We include the sample prompt templates.

\subsubsection{Relative Confidence}
\label{sec:relative-confidence-prompt}
\begin{verbatim}
You are an expert in evaluating questions. 
Compare the following two questions and 
decide which one you are more confident in answering:

Question 1: <Question1>

Question 2: <Question2>

Respond with Question number and reasoning why you are more confident in 
answering a given question. Respond with '<winner>Question 1</winner>
<reason> your reasoning ...</reason>' if you are more confident 
that you can answer Question 1 correctly, or <winner>Question 2 </winner>
<reason> your reasoning ...</reason>' if you are more 
confident in your ability to answer Question 2 correctly.

Answer:
\end{verbatim}

\end{document}